\documentclass[10pt,twocolumn,letterpaper]{article}
\usepackage{wacv}
\usepackage{times}
\usepackage{epsfig}
\usepackage[utf8]{inputenc}
\usepackage{url}
\usepackage{graphicx}
\usepackage{amsmath}
\usepackage{amssymb}
\usepackage{subfigure}
\usepackage{romannum}
\usepackage{caption}
\usepackage{multirow, booktabs, makecell}
\usepackage[table]{xcolor}
\usepackage[accsupp]{axessibility}  

\def\wacvPaperID{716} 

\wacvfinalcopy 

\ifwacvfinal
\def\assignedStartPage{9876} 
\fi

\ifwacvfinal
\usepackage[breaklinks=true,bookmarks=false]{hyperref}
\else
\usepackage[pagebackref=true,breaklinks=true,colorlinks,bookmarks=false]{hyperref}
\fi

\ifwacvfinal
\else
\fi

\begin{document}
\pagenumbering{gobble}
\title{Estimating Image Depth in the Comics Domain
}

\author{Deblina Bhattacharjee, Martin Everaert, Mathieu Salzmann, Sabine Süsstrunk\\
School of Computer and Communication Sciences, EPFL, Switzerland\\
{\tt\small \{deblina.bhattacharjee, martin.everaert, mathieu.salzmann, sabine.susstrunk\}@epfl.ch}
}
\maketitle
\begin{abstract}
   Estimating the depth of comics images is challenging as such images a) are monocular; b) lack ground-truth depth annotations; c) differ across different artistic styles; d) are sparse and noisy. We thus, use an off-the-shelf unsupervised image to image translation method to translate the comics images to natural ones and then use an attention-guided monocular depth estimator to predict their depth. This lets us leverage the depth annotations of existing natural images to train the depth estimator. Furthermore, our model learns to distinguish between text and images in the comics panels to reduce text-based artefacts in the depth estimates. Our method consistently outperforms the existing state-of-the-art approaches across all metrics on both the DCM and eBDtheque images. Finally, we introduce a dataset to evaluate depth prediction on comics.  Our project website can be accessed at \url{https://github.com/IVRL/ComicsDepth}. 
\end{abstract}

\section{Introduction}
\label{sec:introduction}
Depth estimation for comics images can provide important information for applications such as comics image retargeting~\cite{retargeting-depth, rigaud-depth-retargeting}, scene reconstruction~\cite{scene-reconstruction-comics} and reconfiguration of comics~\cite{comics-videogames}, i.e., transferring the stories from paper to an interactive graphical media, for instance, video games based on comics or comics animations. The problem of depth estimation can be framed as that of predicting a metric depth for each pixel in a given input image. In comics, the depth estimation problem is monocular, which makes it inherently ill-posed~\cite{midas}. This is further complicated by the fact that most scenes in the comics domain have large content variations, object occlusions, geometric detailing (different perspectives and size scales), sparse or noisy scenes and non-homogeneous illustrations as shown in Figure 1. As a consequence, while estimating the depth of a comics scene is easy for humans, it remains highly challenging for computational models.

To address this, we explore the extensive research done in the field of monocular depth estimation over the past years, which reports computational models that leverage
monocular cues, such as perspective information, object sizes, object localization, and occlusions, to estimate the depth of scenes~\cite{SIDE}. Note that, while much work has also been done for depth estimation from stereo images~\cite{Bi3D, tankovich2021hitnet,  tonioni2019realtime} or video sequences~\cite{kopf2021rcvd, Luo-VideoDepth-2020}, such approaches do not match the monocular setting we face in the comics domain. 

\begin{figure}[ht]
\centering
\renewcommand{\thesubfigure}{}
\subfigure[Comics Input]
{\includegraphics[ height = 2.5cm, width=3.5 cm ]{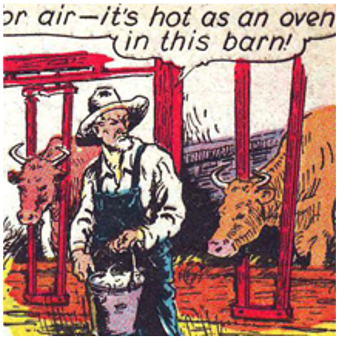}}
\subfigure[CDE~\cite{CDE}- Comics Image]
{\includegraphics[ height = 2.5cm, width=3.5 cm]{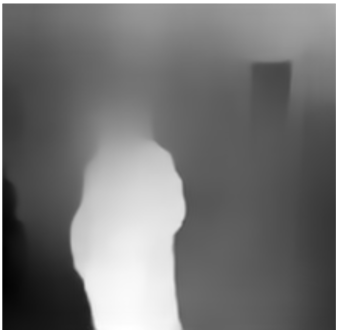}}
\subfigure[Translated Image ]
{\includegraphics[ height = 2.5cm, width=3.5 cm]{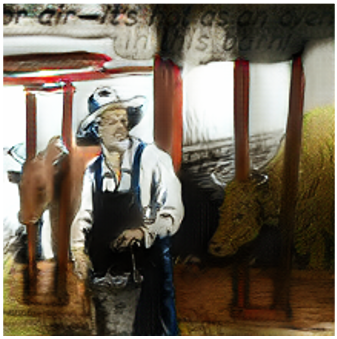}}
\subfigure[CDE~\cite{CDE}- Translated Image]
{\includegraphics[  height = 2.5cm, width=3.5 cm]{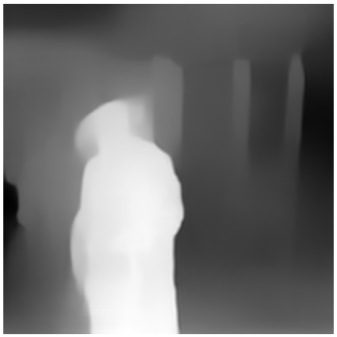}}
\vspace{-11.3 pt}
\caption{{\textbf{Leveraging monocular depth estimation models.} When employed directly on a comics image, the state-of-the-art monocular depth estimation model~\cite{CDE} fails to predict accurate depth. We therefore, translate the comics image to the natural image domain and then apply the CDE depth estimator as mentioned in~\cite{CDE}. } 
}\label{fig:monocular-model-prediction}\vspace{-10pt}
\end{figure}

Because the state-of-the-art monocular depth estimation models~\cite{CDE, midas} have been trained on natural images, they fail to predict the depth of comics images accurately, resulting in vague, overlapping or missing  objects (Figure 1). 
An immediate solution would be to retrain the depth models on comics images, either in a supervised manner, which would in turn require ground-truth depth annotations of comics images, or in an unsupervised manner, which would require employing domain adaptation techniques~\cite{Zheng2018767}. As there exist no dataset with ground-truth depth annotations for comics images and manually annotating the depth of a large number of comics images would be expensive and time-consuming, we employ an unsupervised image-to-image (I2I) translation method~\cite{Bhattacharjee_2020_CVPR} to translate the images from the comics domain to the real one. Once translated to the real domain, we leverage the ground-truth depth of real images to train our depth model and thereby predict the depth of the translated comics$\longrightarrow$real image. The result of this process, compared to the direct application of a trained depth estimation network, is shown in Figure~\ref{fig:monocular-model-prediction}. 

To improve the performance of the depth estimation, we exploit contextual attention, both spatially and channel-wise, as focusing on the scene context parallels how humans estimate the depth of a scene. 
To this end, we introduce a local context model that leverages a Laplacian edge detector to guide depth estimation. This builds on the intuition that depth features significantly depend on edge cues and yields a sharper foreground vs. background separation. Furthermore, we incorporate
a feature-based GAN that encourages the inner feature representations of the depth model to follow similar distributions for the real and translated images.
Additionally, we include a text detector in our model to remove the artefacts in the depth predictions arising from the text or speech balloons in comics images. 

Our main contributions therefore are as follows:
\begin{itemize}
\vspace{-0.2cm}
\item We introduce a cross-domain depth estimation method by leveraging an off-the-shelf unsupervised I2I translation method.  
\vspace{-0.2cm}
\item We exploit the contextual information for depth prediction of a given scene. We use an inner feature-based GAN to enforce similarity between the domains, as well as a Laplacian edge detector to obtain distinct foreground vs. background separations.
\vspace{-0.2cm}
\item By introducing a text detector in our cross-domain depth model, we reduce the artefacts from text and speech balloons in the depth predictions, which are specific to comics.
\vspace{-0.2cm}
\item Finally, we introduce a  benchmark dataset for comics images with 450 manually annotated image-depth pairs comprising 300 images from the standard DCM~\cite{dcm} validation dataset and 150 images from the standard eBDtheque~\cite{eBDtheque} validation dataset. This can be used as a benchmark for future papers for depth evaluations, as there is no existing benchmark with depth annotations for comics.
\end{itemize}
Our experiments on our manually annotated benchmark show that our approach outperforms the state-of-the-art unsupervised monocular depth estimation methods across all the different comics styles. 

\section{Related Work}
\subsection{Monocular Depth Estimation}
Over the past decade, there has been a significant development in monocular depth estimation. Laina et al.~\cite{laina2016deeper} proposed fully convolutional networks with the fast up-projection method using residual learning to model the mapping between RGB images and depth maps. Kuznietsov et al.~\cite{Kuznietsov_2017_CVPR} introduced a semi-supervised approach to overcome the deficiency and limitation of sparse ground-truth lidar maps. Godard et al.~\cite{monodepth17-left-rightconsistency} suggested unsupervised training objective to replace the use of labeled depth maps. The network generates the left and right disparity maps and calculates the reconstruction, smoothness, and left-right consistency losses. Guo et al.~\cite{guo2018learning} incorporated a synthetic depth dataset to acquire a considerable amount of ground-truth images. Subsequently, they trained a network with synthetic data and fine-tuned with a real dataset. Finally, they mitigated the domain gap between the ground-truth and synthetic dataset by distilling stereo networks. Amirkolaee and Arefi~\cite{AMINIAMIRKOLAEE2019105714} constructed a depth prediction network with the encoder–decoder and skip connection structure to integrate the global and local contexts. In~\cite{CDE}, a context based monocular depth estimation method exploits the contextual information between objects via inter object attention to extract visual cues for estimating depth.
While these approaches produce improved and consistent depth results, training them is challenging because of 1) inherently different representations of depth: direct vs. inverse depth representations~\cite{Lore_2018_CVPR_Workshops, Johnston_2020_CVPR}, 2) scale ambiguity: for some data sources, depth is only given up to an unknown scale~\cite{pyramid-stereo-matching, camera-motion-depth-from-video, edge-aware-depth}, 3) shift ambiguity: some datasets provide disparity only up to an unknown global disparity shift~\cite{shift-in-dynamic-scenes}. Further, in the presence of occluded regions (i.e. groups of pixels that are seen in one image but not the other), these methods produce meaningless values due to failed disparity calculations.
To mitigate this, in~\cite{midas}, the authors propose a new loss function that is invariant to both scale and global shift so that the monocular depth estimation model can learn from diverse ground-truth depth maps obtained from disparate domains. Nevertheless, it does not generalise well to either paintings or comics domain. 

With the development of image style transfer and its connection with domain adaptation, researchers adopted the style transfer and adversarial training to estimate depth maps in real scenes~\cite{atapour2018, Kundu_2018_CVPR}, which relied on the models trained with a large amounts of synthetic data. DispNet~\cite{mayer2016} was the first network that introduced image style transfer for depth estimation. Thereafter, Zheng et al.~\cite{Zheng2018767} proposed a two-module domain adaptive network, T2Net, where one module was trained with synthetic and real images and reconstructed each other with the reconstruction loss and generative adversarial loss~\cite{ CHEN2020468, Kumar2018}, and these outputs were input into the other module to predict the real depth maps. As this method is close to our approach, we consider the T2Net as a baseline for comparison. Besides, there are more models with cycle consistency~\cite{zhao2019geometry}, cross-domain~\cite{guo2018learning, Tonioni2020}, and others for domain adaptation to predict monocular depth maps. 
 In this vein, we apply an unsupervised I2I translation method to minimize the domain disparity between comics and real world.
 
\subsection{Domain Adaptation via I2I Translation}
 The advent of I2I translation methods began with the invention of conditional GAN\cite{INIT26}, which have been applied to a multitude of tasks, such as scene translation~\cite{INIT13} and sketch-to-photo translation~\cite{INIT33}. While conditional GANs yield impressive results, they require paired images during training. Unfortunately, in comics$\longrightarrow$real I2I translation scenario, such paired training data is lacking and expensive to collect. To overcome this, cycleGAN~\cite{cyclegan}, with its cycle consistency loss between the source and target domains, is a possible solution for translating the comics images to real images, thereby producing consistent images. Nevertheless, neither conditional GANs, nor cycleGAN account for the multi-modality of comics$\longrightarrow$real I2I translation; in general, a single comics image can be translated to real domain in many different, yet equally realistic ways. This is also due to the different artistic styles present in a single comics domain, which in turn, gives rise to intra-comics domain style variability. Addressing this issue of multi-modality, more recently, MUNIT~\cite{MUNIT} and DRIT~\cite{DRIT} introduced solutions by learning a disentangled representation with a domain-invariant content space and a domain-specific attribute/style space. While effective, all the above-mentioned methods perform image-level translation, without considering the object instances. As such, they tend to yield less realistic results when translating complex scenes with many objects. This is also the task addressed by INIT~\cite{INIT} and DUNIT~\cite{Bhattacharjee_2020_CVPR}. While INIT~\cite{INIT} proposed to define a style bank to translate the instances and the global image separately, DUNIT~\cite{Bhattacharjee_2020_CVPR} proposed to unify the translation of the image and its instances, thus preserving the detailed content of object instances. We, therefore, use DUNIT~\cite{Bhattacharjee_2020_CVPR} as our I2I translation model to translate the comics images to real domain. Once translated, we leverage a depth estimator trained with depth annotations from real images, to ultimately, predict the depth of comics images.

\section{Methodology}
\begin{figure*}[ht]
\centering
{\includegraphics[ width=0.83\linewidth ]{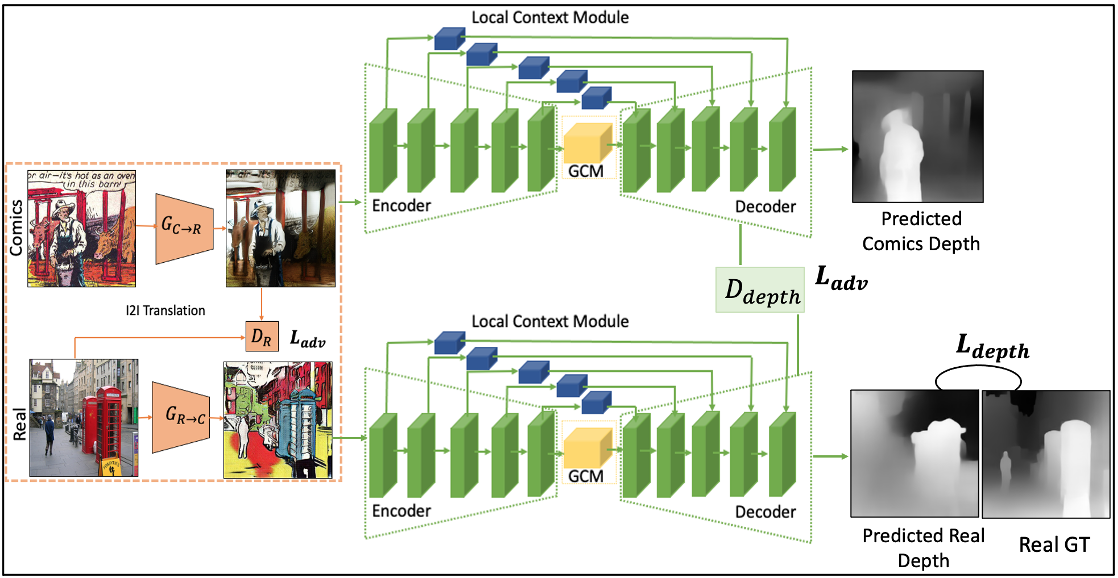}}
{\includegraphics[  height=3.5cm, width=8.3cm]{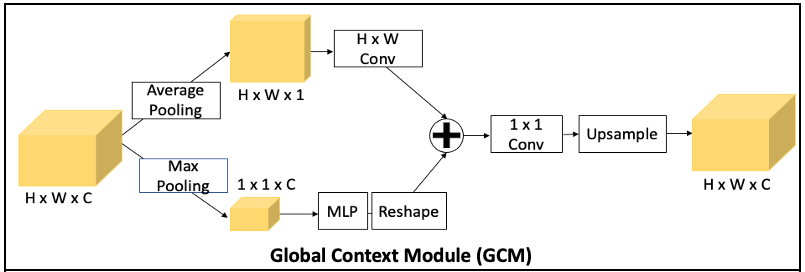}}
{\includegraphics[  height=3.5cm, width=9.0cm]{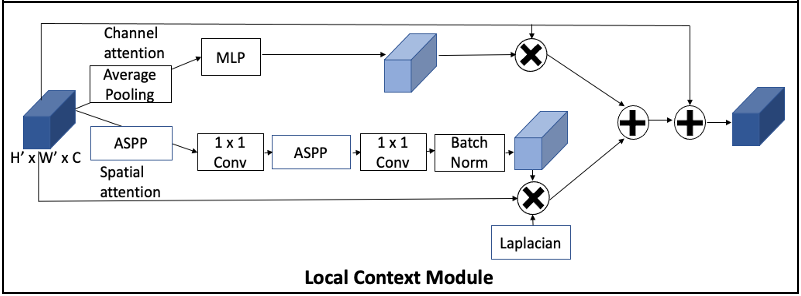}}
\caption{{\textbf{Detailed overview of our architecture.} Top: Overall architecture as discussed in Section 3. Bottom Left: Global Context Module as detailed in~\cite{CDE}. Bottom Right: Local Context Module~\cite{CDE} with the added Laplacian in the spatial attention branch.} 
}\label{fig:detailed-model}\vspace{-10pt}
\end{figure*}
\subsection{Problem Formulation and Overview}
We aim to learn a cross-domain depth mapping between two visual domains $C \subset \mathbb{R}^ {H \times W \times 3}$  and  $R \subset \mathbb{R}^ {H \times W \times 3}$, where $C$ is the comics domain and $R$ is the real image domain. 
To this end, first we employ the DUNIT model~\cite{Bhattacharjee_2020_CVPR} to translate the given comics image to the real domain. Second, we use a contextual monocular depth estimator on the translated image. Thus, the problem can be formulated as
$D_c = f(R(C))$, 
where $D_c$ is the depth prediction for the given comics image $C$, $R(C)$ is the comics$\longrightarrow$real translated image and $f(R(c))$ is the depth estimator trained on real images and applied to $R(c)$. The detailed architecture of our method is provided in Figure~\ref{fig:detailed-model}.
We now explain the components of our network in more detail.

\subsection{Training}
To handle unpaired training images between the comics and real domains, we follow the cycle-consistency approach. In essence, this process mirrors that described in DUNIT~\cite{Bhattacharjee_2020_CVPR}.
Additionally, to study the effect of the I2I translation model on the performance of the depth estimator, we replace the DUNIT method with CycleGAN~\cite{cyclegan} and DRIT~\cite{DRIT}. These methods do not reason about the instance-level translations and thus, perform poorly in contrast to DUNIT. We report these results in the next section.
Below, we detail the loss function and training procedure for the resulting I2I translation based depth model.

\vspace{-10pt}

\paragraph{Image-to-image translation module.}
Our method is built on the DUNIT~\cite{Bhattacharjee_2020_CVPR} backbone which embeds the input images onto a shared style space and a domain specific content space. As such, we use the same weight-sharing strategy as DUNIT for the two style encoders $(E_x^s,E_y^s)$ and exploit the same loss terms. They include:
\begin{itemize}
    \item A content adversarial loss ${\cal L}_{adv}^{content}(E_x^c, E_y^c, D^c)$ relying on a content discriminator ${D^c}$ and the two content encoders $(E_x^c,E_y^c)$, whose goal is to distinguish the content features of both domains;
    \item Domain adversarial losses ${\cal L}_{adv}^{x}(E_y^c,E_x^s,G_x,D^x)$ and ${\cal L}_{adv}^{y}(E_x^c,E_x^{ci},E_y^s,G_y,D^y)$, one for each domain, with corresponding domain classifiers $D^x$ and $D^y$, corresponding domain generators $G_x$ and $G_y$ and instance content encoder $E_x^{ci}$;
    \item A cross-cycle consistency loss ${\cal L}_{1}^{cc}(G_x,G_y,E_x^c,E_x^{ci},E_y^c,E_x^s,E_y^s)$ that exploits the disentangled content and style representations for cyclic reconstruction~\cite{cross-cc}; 
    \item Self-reconstruction losses ${\cal L}_{rec}^x(E_x^c,E_x^{ci},E_x^s,G_x)$ and ${\cal L}_{rec}^y(E_y^c,E_y^s,G_y)$, one for each domain, ensuring that the generators can reconstruct samples from their own domain;
    \item KL losses for each domain ${\cal L}_{KL}^x(E_x^s)$ and ${\cal L}_{KL}^y(E_y^s)$ encouraging the distribution of the style representations to be close to a standard normal distribution;
    \item Latent regression losses ${\cal L}_{lat}^x(E_x^c,E_x^{ci},E_x^s,G_x)$ and ${\cal L}_{lat}^y(E_y^c,E_y^s,G_y)$, one for each domain, encouraging the mappings between the latent style representation and the image to be invertible;
    \item An instance consistency loss ${\cal L}_{1}^{ic}(P_{tl}^{xi},P_{tl}^{yi},P_{br}^{xi},P_{br}^{y})$ encouraging the same object instances to be detected in the source domain image and in the corresponding image after translation, where  $P_{(.)}^{(.)}$ are the bounding box top-left and bottom-right corner pixels for detected instances in the two domains.
\end{itemize}

During training, the I2I module is trained along with the depth estimation module in an end-to-end manner. It has been observed in~\cite{atapour2018, Zheng2018767} that an end-to-end approach yields consistent results on unknown domains, though it comes with a computational overhead. In our method, this computational cost depends mainly on the employed I2I translation module. For instance, DUNIT~\cite{Bhattacharjee_2020_CVPR} has a greater computational overhead than DRIT~\cite{DRIT} or CycleGAN~\cite{cyclegan}. For further details we point the reader to the supplementary material.
\vspace{-10pt}

\paragraph{Depth estimation module.}
As shown in Figure~\ref{fig:detailed-model}, our depth estimation module is an encoder-decoder model with skip connections in between the encoder and decoder layers. These skip connections model the local context in between the visual features by taking into account the spatial and channel attention. The architecture of our depth estimator is inspired from~\cite{CDE}. It relies on a Global Context Module (GCM), mirroring that of~\cite{CDE}, which explores the context of the entire scene, whereby it computes the spatial and channel attention between the objects present in the global image. To this end, the GCM is placed at the end of the encoder to obtain the global context information and pass meaningful features to the decoder. We further complement the GCM with a local context module processing the features extracted at different layers in the encoder of the depth estimator as shown in~\cite{CDE}. Moreover, to clearly contrast the edge boundaries of the objects, we incorporate a Laplacian edge detector~\cite{laplacian} to the spatial branch of the local context module. Since depth leverages low-level visual cues, such as edge information, we have observed this Laplacian to facilitate depth estimation. In particular, the local context module feature (shown in blue in Figure~\ref{fig:detailed-model}), extracted by the encoder of the depth estimator, is processed spatially and channel-wise before being fed into the decoder layer. While the channel-wise processing mirrors that of CDE~\cite{CDE}, the spatial processing (or the spatial attention branch as shown in Figure~\ref{fig:detailed-model}) employs multiple ASPP~\cite{ASPP} and convolutions to obtain a spatially-pooled feature, which is then multiplied with the original local context module feature and the Laplacian~\cite{laplacian}. Finally, the features from both the spatial and channel branch are added to the original feature, to produce the processed local context feature. This feature is fed into the decoder layer.

Our method uses two depth estimators, one taking the real images as input and the other the translated images. We use the zero-shot cross domain MIDAS model~\cite{midas} to generate pseudo ground-truth depth for the real domain. Note, however, that we could use any existing real-image dataset with ground-truth depth annotations, such as KITTI~\cite{kitti} or NYU~\cite{NYU-depth}. However, these datasets are restrictive on the diversity of their scenes, i.e., they are not representative of the extreme scene diversity in comics that contain both indoor and outdoor scenes. Therefore, we use the MIDAS model, which was trained on a collection of five diverse real-world datasets comprising both indoor and outdoor scenes. We generate the pseudo ground-truth only once, before training our depth estimators. 

Nevertheless, MIDAS fails when directly applied on comics images (shown in supplementary Figure 2), hence the need for our cross-domain context aware depth estimators. To train them, we initialize both with the MIDAS weights, setting a low learning rate of $10^{-6}$ to update the weights for 100 epochs with the Adam optimizer and the default hyper-parameters of~\cite{CDE}.
During the training phase, we use a shift and scale-invariant log loss function~\cite{midas} as objective function $L_{depth}$ for the depth estimator in the real domain. It can be expressed as
\vspace{-5pt}
\begin{equation} L_{depth}(y,y^{*}) = \frac {1}{n}\sum _{i}{d_{i}^{2}} - \frac {1}{2{n^{2}}}\left ({\sum _{i}{d_{i}}^{2}}\right)\;,\end{equation}
where $d_i=log(y_i) - log(y_i^*)$, $y$ is the predicted depth, $y^*$ is the pseudo ground-truth depth in the real domain and $n$ is the number of pixels indexed by $i$.

As it learns the depth mappings, the depth estimator in the real domain shares its weight with the estimator in the comics$\longrightarrow$real translated domain. We then add an adversarial loss $L_{adv}$ to train the feature-based GAN between the two depth estimators~\cite{Zheng2018767}, which encourages the inner feature representations of the two depth estimators to share similar distributions, since both stylistically represent real images. This loss is written as
\begin{equation}\begin{aligned} L_{adv}(f, D_{depth}) = E_{f_{c'}\in f_{C}}[log(1-D_{depth}(f_{c'}))]\\ + E_{f_{r'}\in f_{R}}[log(D_{depth}(f_{r'}))]\;,\end{aligned}\end{equation} where $f_C$ and $f_R$ represent the encoded features extracted by the encoder of the depth estimators in the translated domain and the real domain respectively, and $D_{depth}$ is the discriminator of the feature GAN.

Altogether,  we write the overall objective function to train our depth estimators as
\begin{equation}\begin{aligned} L_{obj}(f,D_{depth}) = \alpha_{adv}L_{adv}(f, D_{depth}) + \\ \alpha_{depth}L_{depth}(f)\;.\end{aligned}\end{equation}

\paragraph{Text detection module.} 
When the comics images are translated to the real domain, the translated images comprise text areas or speech balloons, which are in turn unknown to the depth estimator trained on the real domain. This leads to text-based artefacts in the depth results as the depth estimator considers such text areas as objects. Therefore, to control the position of the text areas in the translated images, we train a U-net~\cite{unet} in a supervised manner using the eBDtheque dataset~\cite{eBDtheque}, which contains text/speech balloon annotations. 
We mask the depth maps by multiplying them pixel-wise with the compliment of the text-area mask, before using the L1-loss between the (masked) pseudo ground-truth depth and the depth predictions. 
The detailed architecture for our method with the text detection module is given in the supplementary material. 

\section{Experiments and Results}
To validate our method, we conduct experiments on the following datasets.
\subsection{Datasets}
The main datasets used for this work 
are DCM~\cite{dcm} and eBDtheque~\cite{eBDtheque} for the comics domain and the COCO dataset~\cite{mscoco} for the real-world domain.
The \textbf{DCM} dataset comprises 772 full-page images with multiple comics panel images within. We extract 4470 single panel images from these full-page images using the panel annotations. Note that the panel annotations do not contain depth information. We thus, use these DCM panel images to train the I2I model.
The \textbf{eBDtheque} dataset contains 100  full-page images with multiple comics panel images within. Again, we extract 850 single panel images as before. The eBDtheque dataset contains annotations for speech balloons and text lines, which we  use to train a U-net~\cite{unet} to predict the text areas in a comics image. The detected text areas are then used by our depth model to remove text-based artefacts from the depth predictions.
We employ the \textbf{MS-COCO} dataset~\cite{mscoco}, comprising 5000 real-world images, as real-world domain to train the I2I model. 
\vspace{-8pt}
\paragraph{Benchmark for evaluation.}
\begin{figure}[ht]
\centering
{\includegraphics[height=3.5cm, width=4.0cm]{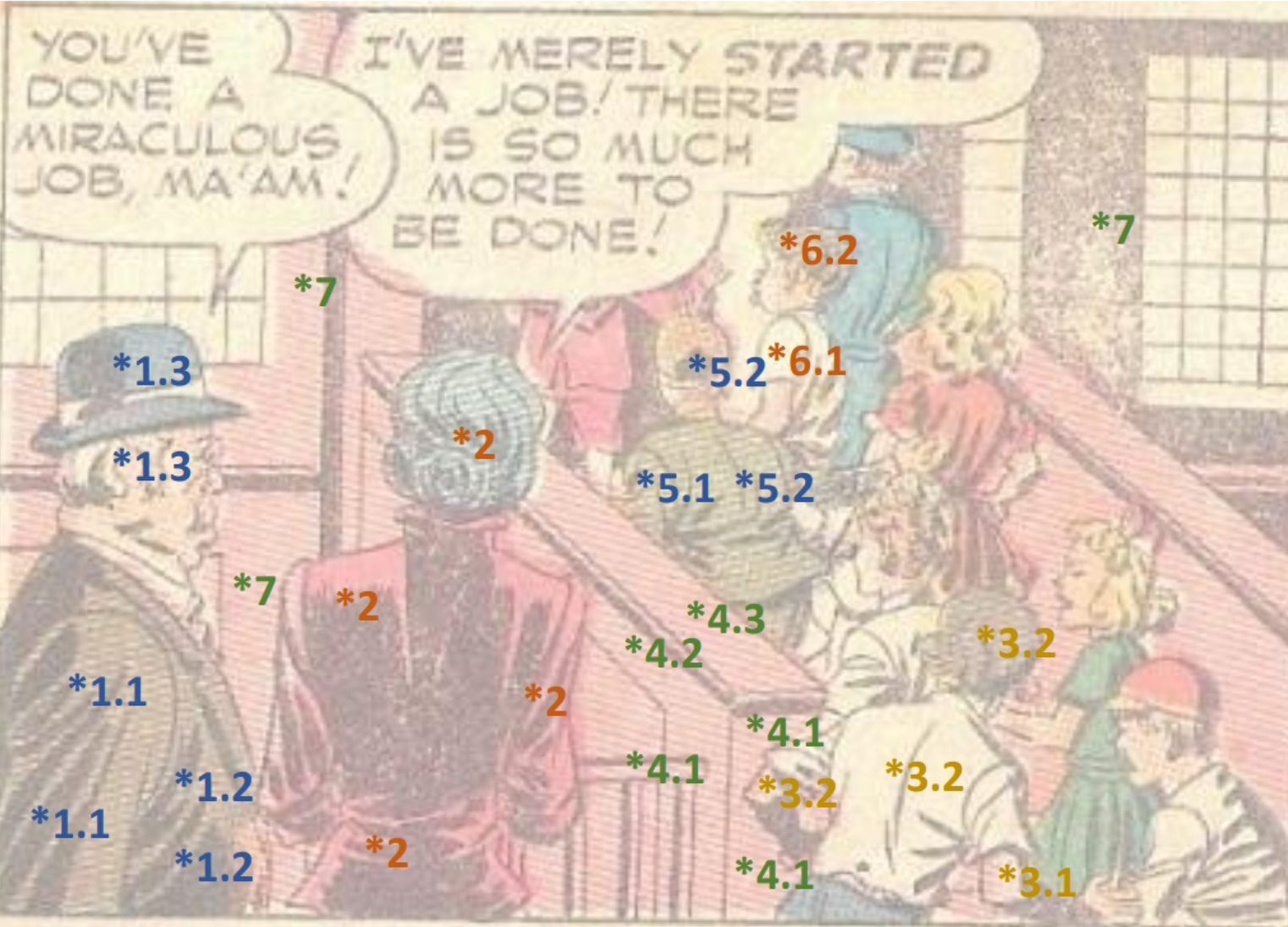}}
{\includegraphics[  height=3.5cm, width=4.0cm]{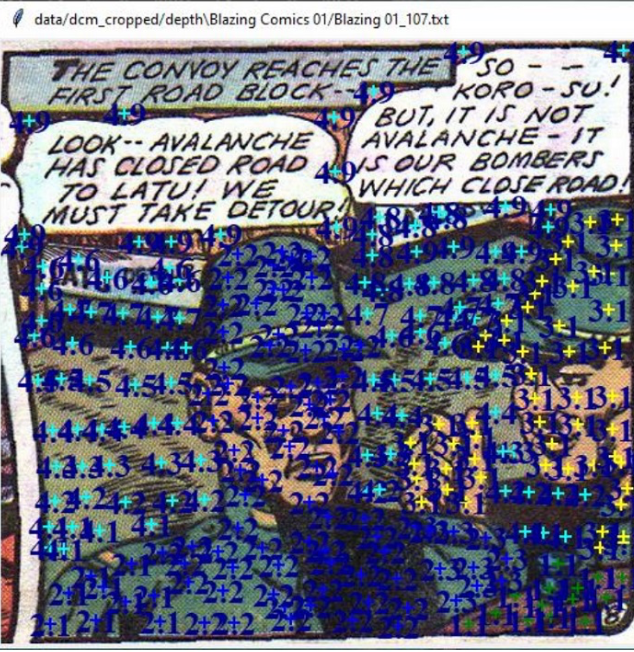}}
\caption{{\textbf{Benchmark for evaluation.} Left: Illustration of the idea of inter-object and intra-object depth ordering, used to annotate the comics images. The closer object is assigned a lower first number $l_1$; and the closer point within the same object is assigned a lower second number $l_2$. Right: Annotated example from the benchmark.} 
}\label{fig:depth-annotations}\vspace{-10pt}
\end{figure}
To evaluate and compare the different depth models, we introduce a benchmark including 300 DCM~\cite{dcm} images and 150 eBDetheque~\cite{eBDtheque} images, from their validation set, along with the corresponding manually annotated ground-truth depth orderings, as illustrated in Figure~\ref{fig:depth-annotations}. To manually annotate their depth, we carefully select 450 images from DCM and eBDtheque validation sets, such that, they contain diverse scenes across ten different artistic styles. These images were further tested for inter-observer variability, for instance, their diversity and artistic styles were analysed by three comics domain experts. Further, all three observers tested the manually annotated depth of comics images. We use depth orderings to annotate the images. In particular, the image pixel coordinates $(x, y)$ are assigned two numbers $(l_1,l_2)$. The first number, $l_1$, represents the inter-object depth ordering, such that two different $l_1$ values imply two different objects.  Closer objects are assigned a lower number. The second number, $l_2$, represents the intra-object depth ordering, such that annotations with the same first number but different second numbers indicate that the two points belong to the same object. A lower $l_2$ value indicates a closer point on the same object.

\subsection{Evaluation Metrics}
To evaluate our method,  we evaluate the following four standard performance metrics, as used in~\cite{CDE, midas}.
\vspace{-10pt}
\paragraph{Absolute relative difference (AbsRel).}
The absolute relative difference is given by $\frac {1}{|N|}\sum _{y\in N}{| y-y^{*}| / y^{*}}$
where $N$ is the number of available pixels in the manually annotated ground-truth.
\vspace{-11pt}
\paragraph{Squared relative difference (SqRel).}
The squared relative difference is defined as $\frac {1}{|N|}\sum _{y\in N}{\|y-y^{*}\|^{2} / y^{*}}$. 
\vspace{-11pt}
\paragraph{Root mean squared error (RMSE).}
The root mean squared error is defined as $\sqrt {\frac {1}{|N|}\sum _{y\in N}{\|y-y^{*}\|^{2}}}$.
\vspace{-11pt}
\paragraph{RMSE (log).} The RMSE (log) is defined as $\sqrt {\frac {1}{|N|}\sum _{y\in N}{\|\log y-\log y^{*}\|^{2}}}$.

\subsection{Quantitative Results}
To evaluate our method, we compare it with the following four state-of-the-art depth estimation approaches.
\begin{itemize}
    \item T2Net~\cite{Zheng2018767}, which comprises a depth prediction model trained on synthetic image-depth pairs.
    \vspace{-0.1cm}
    \item Song et al.~\cite{song-laplacian}, which incorporates a Laplacian pyramid into the decoder architecture. In particular, the encoded features are fed into different streams for decoding depth residuals, defined by the Laplacian pyramid, and the corresponding outputs are progressively combined to reconstruct the final depth map from coarse to fine scales. 
    \vspace{-0.1cm}
    \item MIDAS~\cite{midas}, which introduces a scale and shift-invariant loss to estimate depth from a large collection of mixed real-world datasets, thereby presenting a depth model that generalises across multiple real-world datasets. 
    \vspace{-0.1cm}
    \item CDE~\cite{CDE}, which proposes an architecture that leverages contextual information in a given scene for monocular depth estimation. Thus, using the contextual attention it obtains meaningful semantic features to enhance the performance of the depth model.
\end{itemize}
We report the standard evaluation metrics for our method in comparison with the four state-of-the-art methods in Table~\ref{tb:dcm-depth-results} and Table~\ref{tb:eBDtheque-depth-results} on the DCM and eBDtheque images, respectively. Note that to report the performance metrics, we compare the predicted depth by each method with our manually annotated ground-truth depth. For the results in Table~\ref{tb:dcm-depth-results}, we use the 300 manually annotated DCM image-depth pairs from our benchmark.
Further, for the results in Table~\ref{tb:eBDtheque-depth-results}, we use the 150 manually annotated eBDtheque image-depth pairs from our benchmark. Our method outperforms the baselines on all the performance metrics for both DCM and eBDtheque images. 
Note that to evaluate the performance of the four state-of-the-art methods, the comics image is translated to the real domain using a pretrained DUNIT model and then, the respective methods are applied to predict its depth. This is imperative as the above state-of-the-art methods are trained on real domain, and thus to evaluate them fairly on comics, we translate the comics image to the real domain. To maintain consistency, we also evaluate our approach on the translated comics$\longrightarrow$real image. Nevertheless, our approach can also be directly applied on a comics image to predict its depth. We show this qualitatively in the supplementary material.
\begin{table}[ht]
\setlength\tabcolsep{3.5pt}
\centering
\begin{tabular}{lllll} 
\toprule
Method      & AbsRel$\downarrow$ & SqRel$\downarrow$ & RMSE$\downarrow$  & RMSE log$\downarrow$  \\ 
\midrule
T2Net~\cite{Zheng2018767}       &   0.351     & 0.416      &  1.117     &    0.415       \\
Song et.al.~\cite{song-laplacian} &   0.339     &   0.401    &  1.098     &    0.402       \\
MIDAS~\cite{midas}       & 0.309  & 0.381 & 1.033 & 0.375   \\
CDE~\cite{CDE}       & \underline{0.304}  & \underline{0.374} & \underline{1.024} & \underline{0.367}     \\
Ours        & \textbf{0.251}  & \textbf{0.318} & \textbf{0.971} & \textbf{0.305}     \\
    \bottomrule
    \end{tabular}%
    \setlength{\abovecaptionskip}{1mm}
    \caption{\textbf{Quantitative comparison (DCM images).} We compare our approach with the state-of-the-art methods on the translated DCM validation images~\cite{dcm} from our benchmark. We report the Absolute Relative Difference (AbsRel), Squared Relative Difference (SqRel), Root Mean Squared Error (RMSE), and RMSE log (lower the better). Our contextual depth estimator with the feature-based GAN, Laplacian and text detection module gives the best result. The best results are in bold and the second-best are underlined.} 
   \label{tb:dcm-depth-results}%
\vspace{-10pt}
\end{table}%


\begin{table}[ht]
\setlength\tabcolsep{3.5pt}
\centering
\begin{tabular}{lllll} 
\toprule
Method      & AbsRel$\downarrow$ & SqRel$\downarrow$ & RMSE$\downarrow$  & RMSE log$\downarrow$  \\ 
\midrule
T2Net~\cite{Zheng2018767}       &  0.491     &  0.555     &   1.459    &  0.777        \\
Song et.al.~\cite{song-laplacian} &  0.479     &   0.520    &  1.431     &  0.711       \\
MIDAS~\cite{midas}       & \underline{0.419}  & \underline{0.503} & 1.416 & 0.659    \\
CDE~\cite{CDE}       & 0.424  & 0.511 & \underline{1.415} & \underline{0.647}     \\
Ours        & \textbf{0.376}  & \textbf{0.448} & \textbf{1.364} & \textbf{0.553}     \\
    \bottomrule
    \end{tabular}%
    \setlength{\abovecaptionskip}{1mm}
    \caption{\textbf{Quantitative comparison (eBDtheque images).} We compare our approach with the state-of-the-art methods on the translated eBDtheque validation images~\cite{eBDtheque} from our benchmark. We report the Absolute Relative Difference (AbsRel), Squared Relative Difference (SqRel), Root Mean Squared Error (RMSE), and RMSE log (lower the better). Our contextual depth estimator with the feature-based GAN, Laplacian and text detection module gives the best result. The best results are in bold and the second-best are underlined.} 
   \label{tb:eBDtheque-depth-results}%
\vspace{-10pt}
\end{table}%

\subsection{Qualitative Results}
\begin{figure}[ht]
\centering
\renewcommand{\thesubfigure}{}
\subfigure[]
{\includegraphics[width=0.24\linewidth]{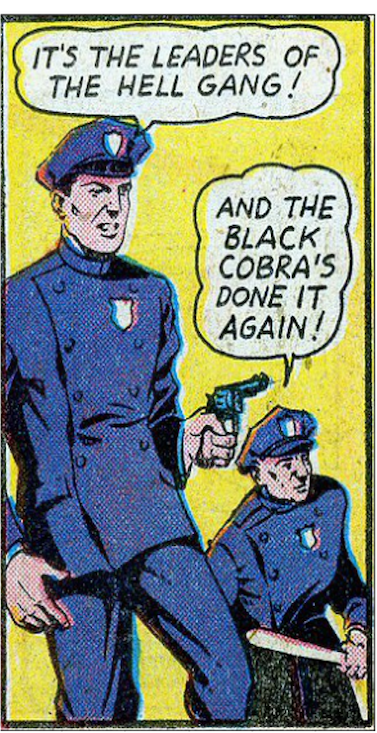}}
\subfigure[]
{\includegraphics[width=0.24\linewidth]{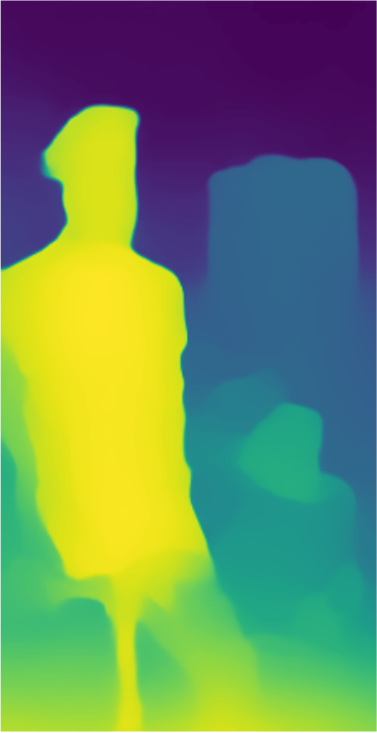}}
\subfigure[]
{\includegraphics[width=0.24\linewidth]{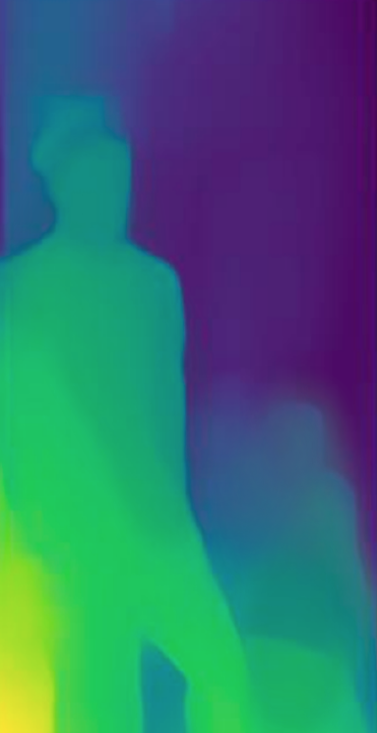}}
\subfigure[]
{\includegraphics[width=0.24\linewidth]{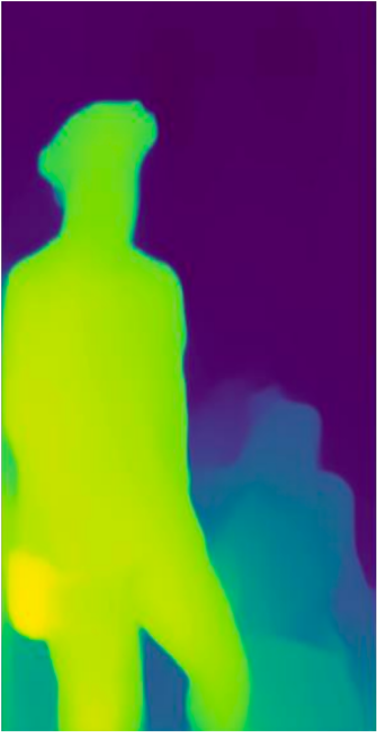}}
\\
\vspace{-21.5pt}
\subfigure[]
{\includegraphics[width=0.24\linewidth]{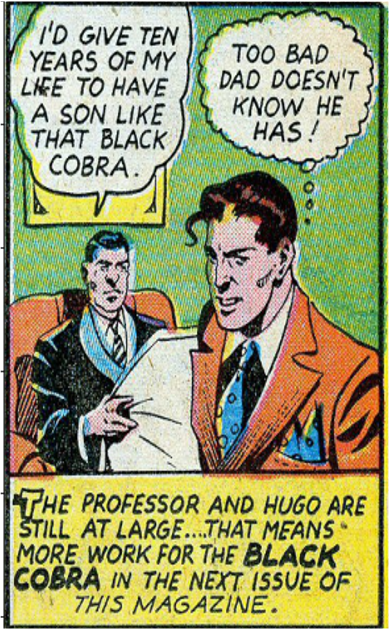}}
\subfigure[]
{\includegraphics[width=0.24\linewidth]{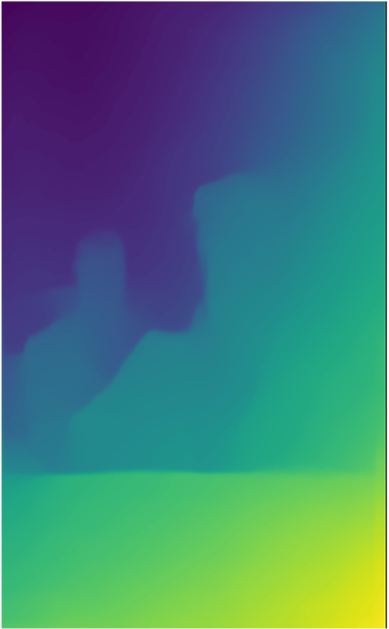}}
\subfigure[]
{\includegraphics[width=0.24\linewidth]{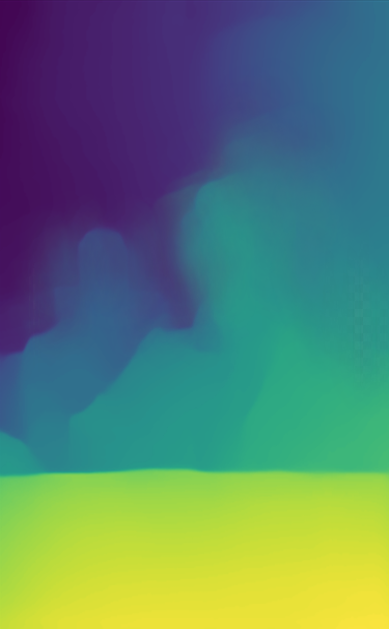}}
\subfigure[]
{\includegraphics[width=0.24\linewidth]{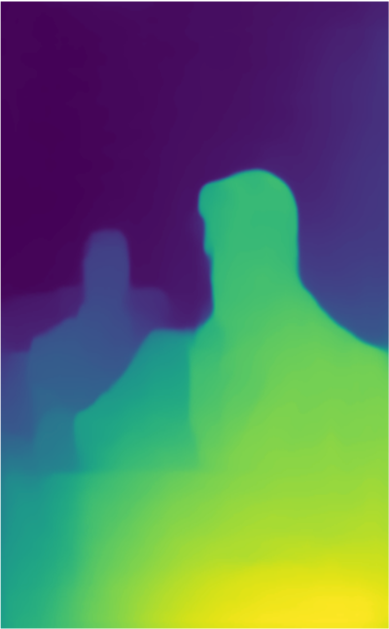}}
\\
\vspace{-21.5pt}
\subfigure[Comics input]
{\includegraphics[width=0.24\linewidth]{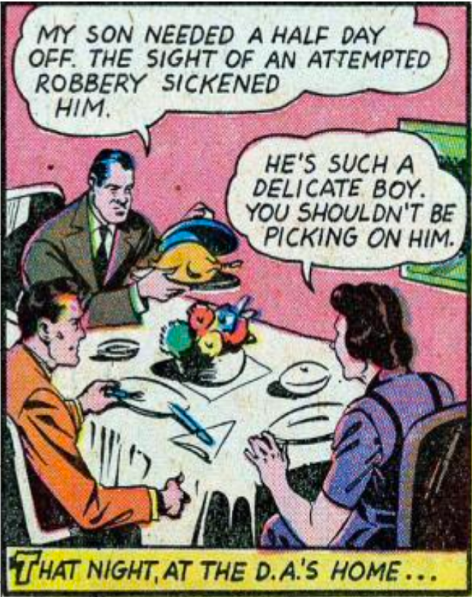}}
\subfigure[MIDAS~\cite{midas}]
{\includegraphics[width=0.24\linewidth]{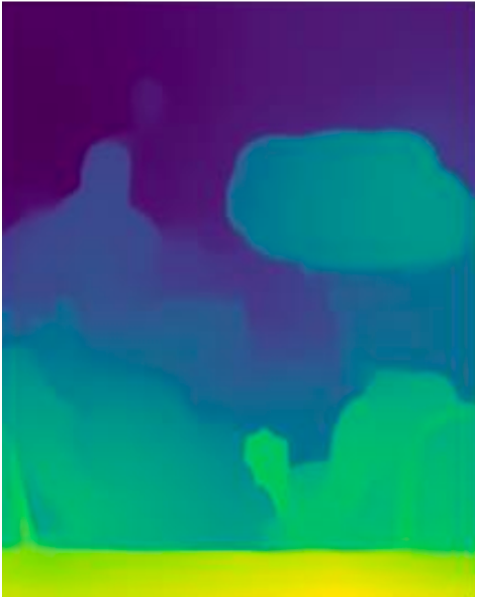}}
\subfigure[CDE~\cite{CDE}]
{\includegraphics[width=0.24\linewidth]{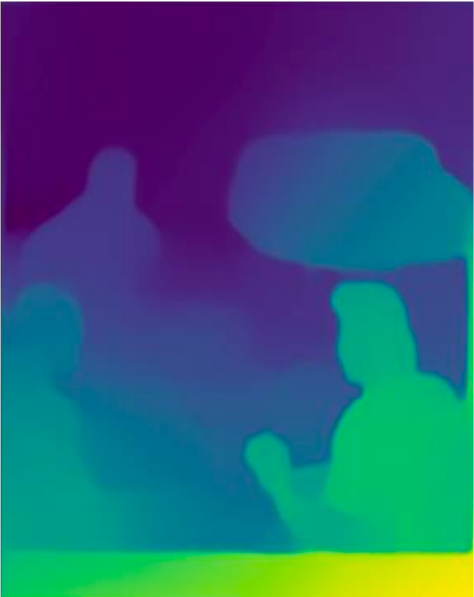}}
\subfigure[Our]
{\includegraphics[width=0.24\linewidth]{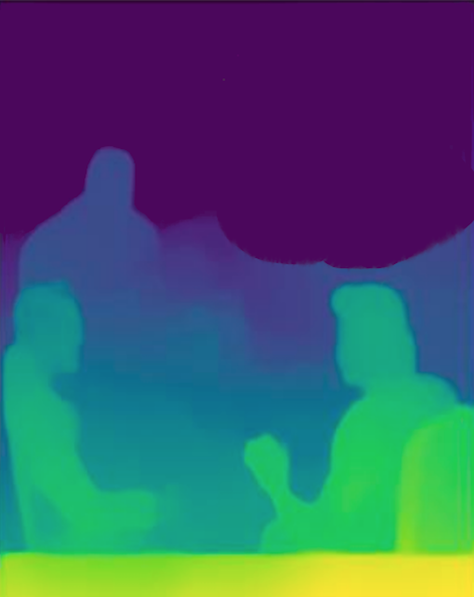}}
\vspace{-10pt}
\caption{\textbf{Qualitative comparison of depth estimation} on the translated DCM validation images~\cite{dcm} from our benchmark, using the text detection module (top row and middle row) and without using the text detection module (bottom row). We show, from left to right, the input image in the comics domain, the result using the MIDAS~\cite{midas} model directly on the translated comics image, the result using the CDE~\cite{CDE} model directly on the translated comics image, and \textbf{Our} model applied to the translated comics image, respectively.
}\label{fig:qualitative-result}\vspace{-10pt}
\end{figure}

In Figure~\ref{fig:qualitative-result}, we compare our method with the depth predictions obtained by MIDAS~\cite{midas} and CDE~\cite{CDE}. The examples demonstrate that our network can benefit from I2I translation in addition to the feature-based GAN and Laplacian. Moreover, we also qualitatively show the effect of our text-detection module. For instance, in the middle row of Figure~\ref{fig:qualitative-result}, while MIDAS and CDE have text-based artefacts in the predictions, including vague depth values in the background from the speech balloons and incorrect depth from the text box in the foreground, our method correctly removes the speech balloon artefacts. Further, our model predicts the human object in the same depth plane as that of the text box in the foreground. Note that these predictions were verified by comics domain experts. 
Our method threfore, yields sharper depth maps with clearer foreground vs. background separation and with well-defined object edges. Furthermore, in contrast to the baselines, the depth predictions by our method show greater consistency in their intra-object and inter-object depth values. 

\subsection{Ablation Study}
We now evaluate different aspects of our method. First, we study the influence of the I2I translation module on our depth model (including the feature GAN, Laplacian and the text module). To this end, we compare the results obtained using our depth model with the different state-of-the-art I2I method, namely, cycleGAN~\cite{cyclegan}, DRIT~\cite{DRIT} and DUNIT~\cite{Bhattacharjee_2020_CVPR}. We report the  AbsRel, SqRel, RMSE and RMSE (log) on the DCM validation images~\cite{dcm} from our benchmark in Table~\ref{tb:effect-of-I2I}. We observe that DUNIT consistently improves the results across all metrics, thereby demonstrating the benefits of instance-level translations on our method, in contrast to the image-level translations in cycleGAN and DRIT.

\begin{table}[ht]
\setlength\tabcolsep{3.3pt}
\centering
\begin{tabular}{lllll} 
\toprule
Method      & AbsRel$\downarrow$ & SqRel$\downarrow$ & RMSE$\downarrow$  & RMSE log$\downarrow$  \\ 
\midrule
CycleGAN~\cite{cyclegan} &   0.282    &  0.346    &  0.995    &   0.329       \\
DRIT~\cite{DRIT} &  \underline{0.269}     & \underline{0.333}     & \underline{0.983}     &  \underline{0.317}      \\
DUNIT~\cite{Bhattacharjee_2020_CVPR}       & \textbf{0.251}  & \textbf{0.318}  & \textbf{0.971} & \textbf{0.305}  \\
    \bottomrule
    \end{tabular}%
    \setlength{\abovecaptionskip}{1mm}
    \caption{\textbf{Ablation Study on the effect of I2I model.} We compare the effect of the different I2I translation model on our method. We report the four standard performance metrics (lower the better). Our method with the DUNIT model gives the best result. The best results are in bold and the second-best are underlined. Note that the DCM validation images~\cite{dcm} from our annotated benchmark were used for this ablation study.} 
   \label{tb:effect-of-I2I}%
\vspace{-10pt}
\end{table}%

We then turn to exploring the effect of the feature GAN, Laplacian and text detection module on our method. To this end, we add each of these components one-by-one to the baseline approach comprising the DUNIT model and the CDE model, shown as I2I$+$depth in Table~\ref{tb:ablation}. Note that this baseline approach is trained in an end-to-end manner. We report the standard four performance metrics on the DCM~\cite{dcm} images from our benchmark in Table~\ref{tb:ablation}. We show that the end-to-end baseline approach outperforms the CDE~\cite{CDE} method when applied directly to the translated comics images, as shown in Table~\ref{tb:dcm-depth-results}. This solidifies the benefits of an end-to-end training approach. Moreover, the addition of each component of our method consistently improves the performance across all metrics. All the images were kept constant for the study of all the network components. We show qualitative results from this ablation study in our supplementary material.

\begin{table}[ht]
\setlength\tabcolsep{3.3pt}
\centering
\begin{tabular}{lllll} 
\toprule
Method      & AbsRel$\downarrow$ & SqRel$\downarrow$ & RMSE$\downarrow$  & RMSE log$\downarrow$  \\ 
\midrule
I2I + Depth       &   0.301   &   0.369   & 1.022     &  0.362     \\
Feature GAN       &  0.270     &   0.339   &  0.994     &     0.322     \\
Laplacian         &  \underline{0.257}    &  \underline{0.322}     &  \underline{0.976} & \underline{0.313}   \\
Text Module      & \textbf{0.251}  & \textbf{0.318} & \textbf{0.971} & \textbf{0.305}   \\
    \bottomrule
    \end{tabular}%
    \setlength{\abovecaptionskip}{1mm}
    \caption{\textbf{Ablation Study on the effect of the different network components.} We compare the effect of the different network components, namely, the feature GAN, Laplacian and text module on our method. We report the four standard performance metrics (lower the better). The above network components are added one-by-one and we observe that \emph{our model with feature GAN, Laplacian and text module} outperforms on all performance metrics. The best results are in bold and the second-best are underlined. Note that the DCM images from our benchmark were used for this ablation study.} 
   \label{tb:ablation}%
\vspace{-15pt}
\end{table}%
\section{Conclusion}
We have introduced an approach to estimate image depth in the comics domain using unsupervised I2I translation to adapt the comics images to the real domain. To this end, we have leveraged a modified context-based depth model trained on real-world images with Laplacian. We also, have added a feature GAN approach to the depth estimators to enforce the semantic similarity between the translated and real images. We have further added a text-detection module to remove text-based artefacts in the depth predictions. To validate our experiments, we introduce a benchmark with manually annotated depth for images from the validation set of DCM and eBDtheque datasets, as there is no existing benchmark with depth annotations. In our experiments, our I2I translation-based modified depth estimators with Laplacian, feature GAN and text-detections, outperform the state-of-the-art methods. This is the first automated method to predict depth for comics images. Therefore, this work can be used for applications like comics image retargeting, scene reconstruction, comics animations or repurposing comics to augmented reality. 
\paragraph{Acknowledgement.}
This work was supported in part by the Swiss National Science Foundation via the Sinergia grant CRSII5$-$180359.
{\small
\bibliographystyle{ieee_fullname}

}

\end{document}


\pagenumbering{gobble}
\title{Estimating Image Depth in the Comics Domain (Supplementary)
}

\author{Deblina Bhattacharjee, Martin Everaert, Mathieu Salzmann, Sabine Süsstrunk\\
School of Computer and Communication Sciences, EPFL, Switzerland\\
{\tt\small \{deblina.bhattacharjee, martin.everaert, mathieu.salzmann, sabine.susstrunk\}@epfl.ch}
}
\maketitle

In this supplementary material, we provide details about the text-detection module, additional qualitative comparison for the state-of-the-art methods with our approach, qualitative results for the ablation study of our network and an analysis on the computational cost of our network components. The document is structured as follows:
\begin{itemize}
    \item Section \textbf{\textcolor{red}{1}}: Text-detection Module
    \item Section \textbf{\textcolor{red}{2}}: Qualitative Comparison- Depth Results
    \item Section \textbf{\textcolor{red}{3}}: Qualitative Results- Ablation Study
    \item Section \textbf{\textcolor{red}{4}}: Computational Cost Analysis
\end{itemize}

\section{Text-detection Module} 
\begin{figure*}[hbt]
\centering
\renewcommand{\thesubfigure}{}
\subfigure[\textbf{Motivation for our text module}]
{\includegraphics[width=0.8\linewidth]{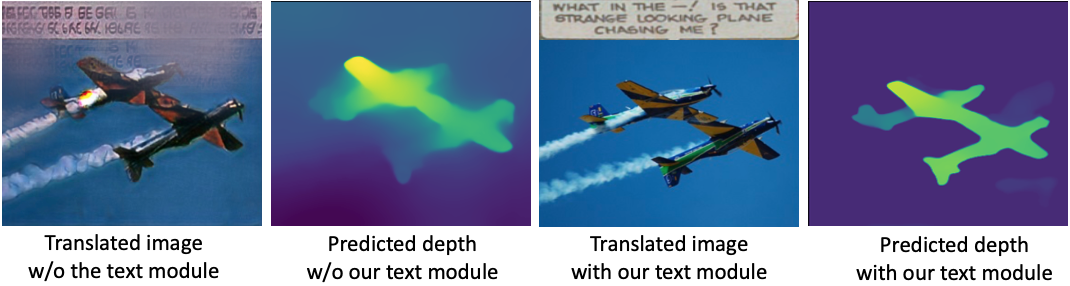}}\\
\subfigure[\textbf{Overview of our approach with the text module}]
{\includegraphics[width=0.9\linewidth]{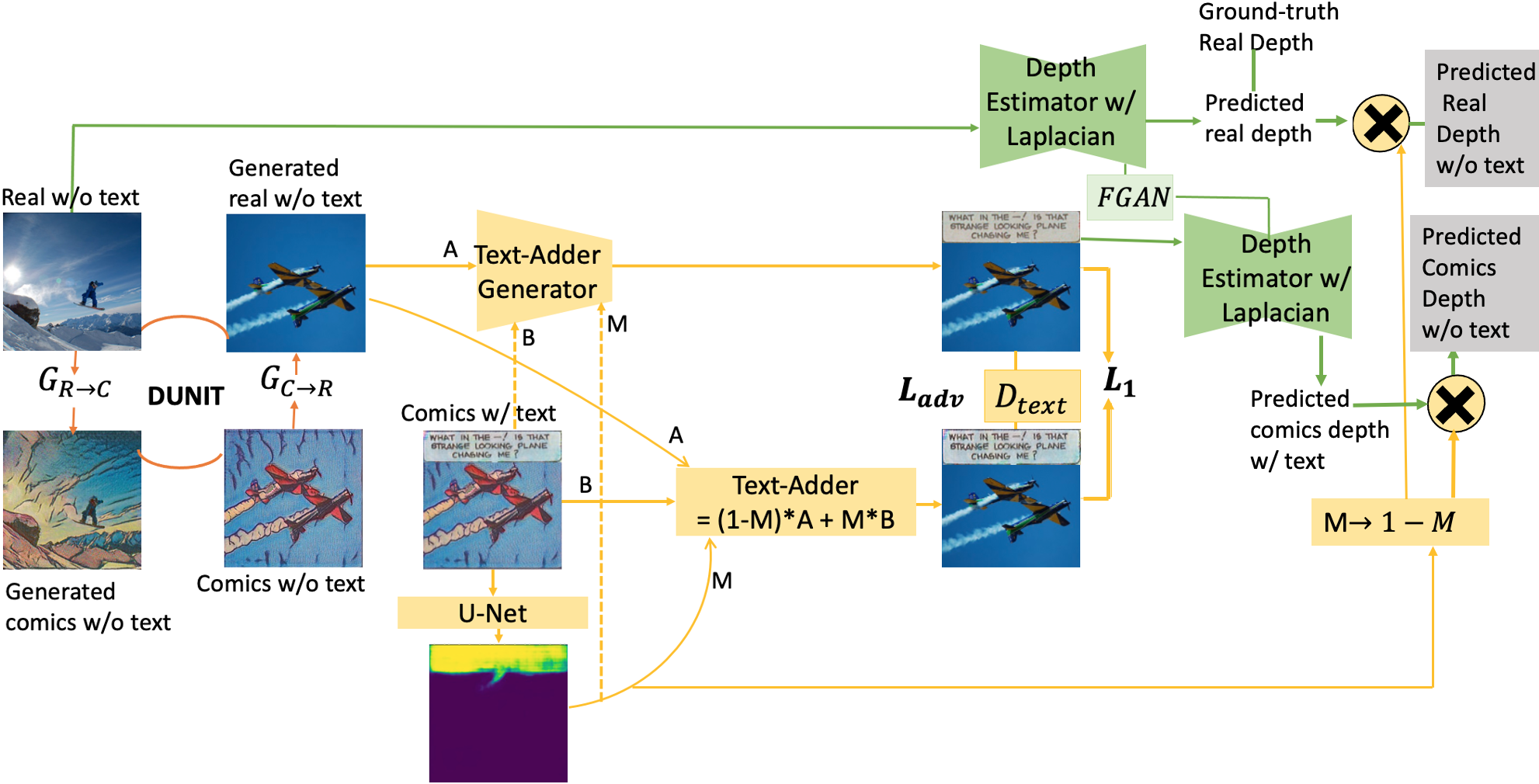}}\\
\vspace{-10pt}
\caption{\textbf{Our depth estimation approach with the text-detection module.} We show, (Top): the motivation for our text module and (Bottom): the overall architecture of our approach incorporating the text module. 
In our text module, the masks are generated using a U-Net~\cite{unet} trained on the text annotations form eBDtheque~\cite{eBDtheque} dataset. The generated masks by the trained U-Net is used to train the text adder generator and the text adder 'ground-truth' as discussed above. The generated real image with text is then fed into the depth estimator. This predicts the depth with text. To remove the text based artifacts in the depth prediction, the complement of the text mask is multiplied with the predicted depth with text to, finally, predict the depth without the text.   
}\label{fig:text-module-architecture}\vspace{-10pt}
\end{figure*}

The generated real-images from the DUNIT~\cite{Bhattacharjee_2020_CVPR} model have speech-balloons or text present in them, which are not recognised by the depth estimators trained on real-domain images. Therefore, the predicted depths contain text-based artefacts. In order to remove these artefacts, we use the text-detection module shown in Figure~\ref{fig:text-module-architecture}. Our text-detection module is a U-Net~\cite{unet}, trained in a supervised manner, on the text/ speech-balloon annotations from the eBDtheque~\cite{eBDtheque} dataset. The trained U-Net~\cite{unet} is then, used on the DCM~\cite{dcm} training images to detect the text/ speech-balloon areas in them, in the form of text masks. These text masks are then used to generate the text adder 'ground-truth' given by $(1-M)A+MB$, where M is the text mask, $(1-M)$ is its complement, A is the comics$\longrightarrow$real translated image (but without the text area) and B is the original comics image (containing the text area). Once the text adder 'ground-truth' is created, we train a text-adder generator with $A, B$ and $M$ as input. This generator takes the position of the mask, $M$, in the original comics image, $B$, and applies this positional information onto the comics$\longrightarrow$real translated image, $A$, to create a well-defined text area on the translated image. This generated output is trained using an $L1$ loss with the text adder 'ground-truth'. The reason to create a translated image with a well-defined text area is shown in Figure~\ref{fig:text-module-architecture}, top row, where we can see that a translated image when generated without the text area information contains text-based artefacts, which in turn, gives incorrect depth values after being fed into the depth estimator. However, the text-adder generator output produces no such text-based artefacts and gives a better depth prediction. 

After the translated image with a well-defined text area is created, its fed into our depth estimator to predict the depth of the translated image with the text. Concurrently, the real image is passed to the other depth estimator to predict the depth of the real image. Both these estimators are trained in an end-to-end manner. Furthermore, to predict the depth of the translated image without the depth values from the text masks, we multiply the complement of the text mask with the prediction. This results in a clean depth prediction without any text-based artefacts. During inference, our approach can be directly applied on the original comics image with text. However, for fair comparison with the baseline approaches, we translate the comics image with text to a real image using a pretrained DUNIT (without the text-detection module) and then apply the different methods to predict their depth. As our approach has been trained with text information, it learns to separate the text-based artefacts and thus, produces a superior depth map. In Figure~\ref{fig:qualitative-ablation-supp}, last column, we observe the effect of our text module on the depth predictions for an input comics image from the DCM validation set of our benchmark (Please zoom in to observe the differences in the depth predictions).

Note that for our final approach (consisting I2I, depth, feature GAN, Laplacian and the text module), we use comics images without text areas to train our I2I module. This is done to facilitate the generation of real images without text artefacts (referred to as 'A' in Figure~\ref{fig:text-module-architecture}). To this end, we discuss the method to generate the original comics images without the text areas, in what follows. 
\paragraph{Generating the comics-without-text dataset.}
To remove the text areas from the original comics, we randomly crop the original images along with their respective text mask prediction obtained by the trained U-Net~\cite{unet}, to a 384 x 384 size. We then, decrease the crop size by 1 unit per dimension, i.e., the image is cropped to 383 x 383, followed by 382 x 382, and so on. We repeat this process until the maximum area of the text in the image is 3\% of the total image. After cropping, these images were checked manually for any remaining text areas and we found that none of the images contained significant text in them.

\section{Qualitative Comparison- Depth Results} 
\begin{figure}[hbt]
\centering
\renewcommand{\thesubfigure}{}
\subfigure[]
{\includegraphics[width=0.24\linewidth]{images/result-input-1.png}}
\subfigure[]
{\includegraphics[width=0.24\linewidth]{images/result-MIDAS-1.png}}
\subfigure[]
{\includegraphics[width=0.24\linewidth]{images/result-CDE-1.png}}
\subfigure[]
{\includegraphics[width=0.24\linewidth]{images/results-Ours-1.png}}
\\
\vspace{-21.5pt}
\subfigure[Comics input]
{\includegraphics[width=0.24\linewidth]{images/result-input-1.png}}
\subfigure[MIDAS~\cite{midas}]
{\includegraphics[width=0.24\linewidth]{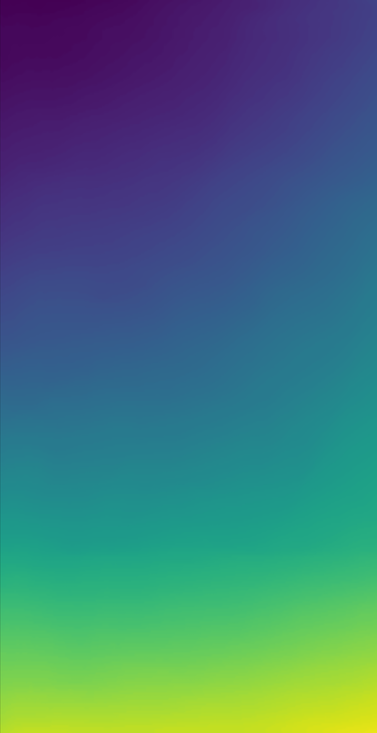}}
\subfigure[CDE~\cite{CDE}]
{\includegraphics[width=0.24\linewidth]{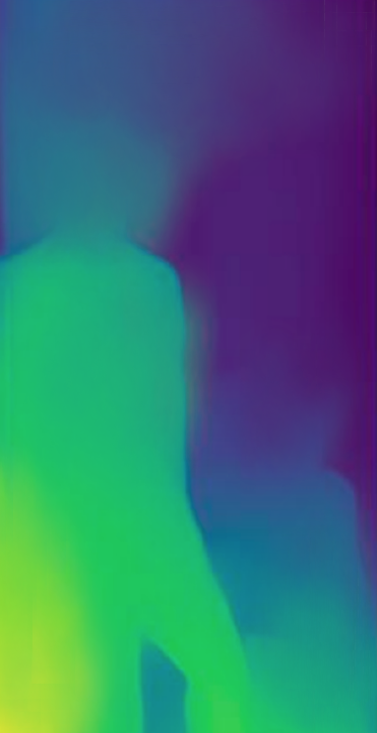}}
\subfigure[Our]
{\includegraphics[width=0.24\linewidth]{images/results-Ours-1.png}}
\vspace{-10pt}
\caption{\textbf{Qualitative comparison of depth estimation} on the DCM validation images~\cite{dcm} from our benchmark. (Top Row): Depth predictions on the translated comics images as seen in the main paper. (Bottom Row): Depth predictions on the actual comics images (not translated). We show, from left to right, the input image in the comics domain, the result using the MIDAS~\cite{midas} model, the result using the CDE~\cite{CDE} model, and \textbf{Our} model (comprising I2I, depth, feature GAN, Laplacian and the text module), respectively. We show that all the methods benefit from the I2I module. Further, we show that our approach can predict depth when applied both to the translated image, as well as the original comics image; while outperforming the baselines in both the scenarios. Cooler colors are farther and warmer colors are nearer (Best viewed in color). 
}\label{fig:qualitative-result-supp}\vspace{-10pt}
\end{figure}

We show the depth predictions on the translated comics images as reported in Figure 5 of the main paper. Further, we show that the state-of-the-art methods like MIDAS~\cite{midas} and CDE~\cite{CDE}, which are trained on real-world images, fail to predict depth accurately when applied to comics images directly. Specifically, as seen in Figure~\ref{fig:qualitative-result-supp}, MIDAS is unable to predict the depth of the sample DCM~\cite{dcm} validation image from our benchmark, though it is trained on a large collection of real-world images from five different real-world datasets. This raises the need for applying these methods on a comics$\longrightarrow$real translated image. As seen in Figure~\ref{fig:qualitative-result-supp}, the baseline methods of MIDAS and CDE (trained on real images), benefit from the I2I translations. During inference, we first, translate the original comics image to the real domain by using a pretrained DUNIT~\cite{Bhattacharjee_2020_CVPR} model and then, we apply the baseline depth estimators on these translated images. Nevertheless, our approach can predict the depth on both the translated image and the original comics image, while outperforming all the baselines in both the scenarios. This is because our approach is trained in an end-to-end manner along with the I2I module. 
Note that, we could also train all the baseline models (including T2Net~\cite{Zheng2018767}, Song et.al~\cite{song-laplacian} and MIDAS~\cite{midas}), from scratch, in an end-to-end manner with our I2I model, but this was not addressed. This is because we observed that training the baseline method of CDE~\cite{CDE} from scratch, in an end-to-end manner, results in poorer results than our approach, as shown in Table 4 (first row) in our main paper. Note that though CDE was the best performing baseline method, it fails in comparison to our approach.  

\section{Qualitative Results- Ablation Study}
\begin{figure}[hbt]
\centering
\renewcommand{\thesubfigure}{}
\subfigure[I2I+D]
{\includegraphics[width=0.24\linewidth]{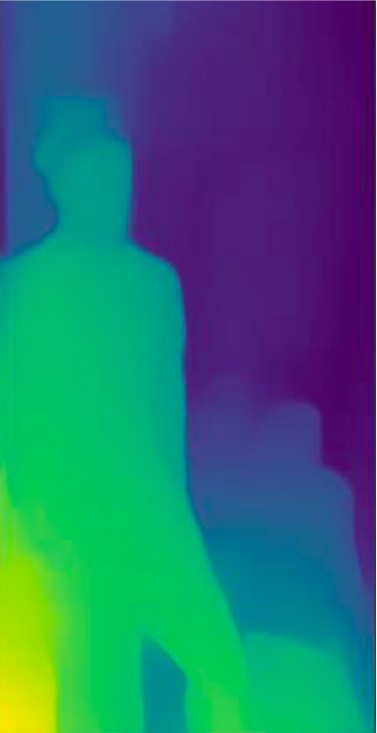}}
\subfigure[I2I+D+FG]
{\includegraphics[width=0.24\linewidth]{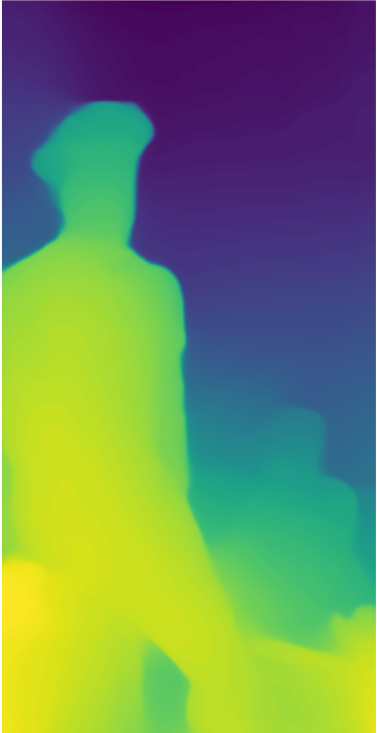}}
\subfigure[I2I+D+FG+L]
{\includegraphics[width=0.24\linewidth]{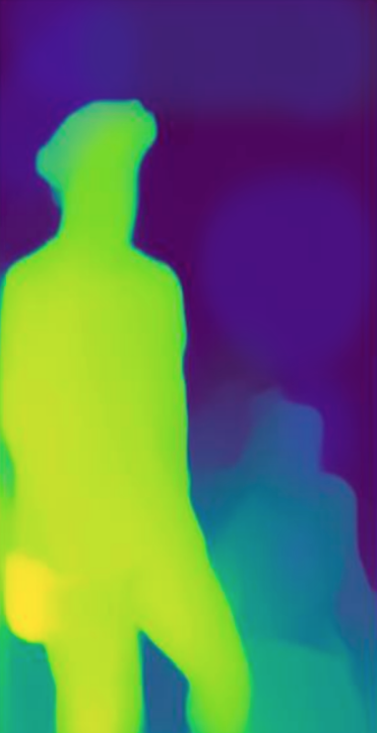}}
\subfigure[I2I+D+FG+L+TM \textbf{(Ours)}]
{\includegraphics[width=0.24\linewidth]{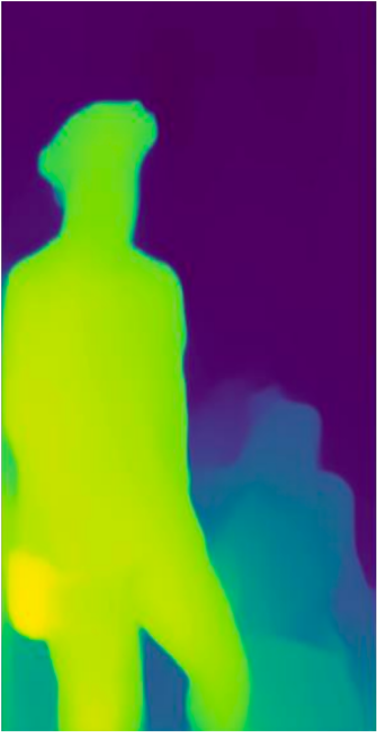}}
\vspace{-10pt}
\caption{\textbf{Qualitative comparison for the ablation study showing the effect of the different network components.} We show the depth predictions on the translated DCM validation images~\cite{dcm} from our benchmark. We report, from left to right, the depth predictions obtained by the model comprising I2I (DUNIT~\cite{Bhattacharjee_2020_CVPR}) and Depth (CDE~\cite{CDE}) trained in an end-to-end manner; the result using the model comprising I2I, CDE and feature GAN; the result using the model comprising I2I, CDE, feature GAN and Laplacian; and \textbf{Our} model (comprising I2I, CDE, feature GAN, Laplacian and the text module), respectively. Cooler colors are farther and warmer colors are nearer (Best viewed in color).
}\label{fig:qualitative-ablation-supp}\vspace{-10pt}
\end{figure}

We validate the results observed in Table 4 of the main paper with additional qualitative results. In Figure~\ref{fig:qualitative-ablation-supp}, we observe the effect of each of our network components on the depth predictions when applied to a translated comics image from the DCM validation set of our benchmark. Note that each network component was added one-by-one. We see, qualitatively, that the DUNIT (I2I)+ CDE (D) method, when trained in an end-to-end manner, outperforms the baseline CDE~\cite{CDE} method (cross-referring to Figure~\ref{fig:qualitative-result-supp}- first row and third column). We also see that the addition of the feature-based GAN (FG) greatly benefits the depth predictions as it encourages the similarity in distribution between the comics and the real domain. Moreover, the Laplacian (L) when added to our depth estimator, refines the edge contrasts and gives a better depth prediction. However, some text-based artefacts still remain in the depth prediction, resulting in vague depth values. To remedy this, we add the text module (TM) to finally, have superior depth predictions as seen in Figure~\ref{fig:qualitative-ablation-supp} here and Table 4 in our main paper.

\section{Computational Cost Analysis}
We have seen, thus far, that training our approach in an end-to-end manner improves the predicted depth maps and thereby benefits our method. However, this leads to a computational overhead. We report the computational cost incurred by the different components of our network, when trained in an end-to-end manner in Table~\ref{tb:computational-cost}. We see that the training of the I2I module dominates the computational time, regardless of the I2I method employed. This was followed by the training of the depth estimators. Note that the training in both the scenarios (i.e. using DUNIT and DRIT as the I2I module) was done using 4 V100, 7 Tflops GPUs with 32 GB memory. The total time taken by our approach with DUNIT~\cite{Bhattacharjee_2020_CVPR} is 36 hours, while that with DRIT~\cite{DRIT} is 27 hours. The extra computational time for DUNIT comes from the instance-level translations. Nevertheless, the inference time for both the methods are comparable and is equal to 217 milliseconds for the method employing DUNIT and 203 milliseconds for the one with DRIT, processed on a single V100 GPU.
\paragraph{Acknowledgement.}
This work was supported in part by the Swiss National Science Foundation via the Sinergia grant CRSII5$-$180359.
\begin{table}
\setlength\tabcolsep{3.3pt}
\centering
\begin{tabular}{lllll} 
\toprule
Method      & w/ DUNIT~\cite{Bhattacharjee_2020_CVPR} & w/ DRIT~\cite{DRIT}  \\ 
\midrule
I2I      &   66\%  &   60\%      \\
Depth (D)      &   17\%  &   17\%      \\
Feature GAN (FG)       &  4\%     &   10.33\%     \\
Laplacian (L)        &  1\%    &  1.33\%    \\
Text Module (TM)        &  12\%    &  11.34\%    \\
\midrule
\textbf{Total}    & \textbf{36h}  & \textbf{27h}   \\
    \bottomrule
    \end{tabular}%
    \setlength{\abovecaptionskip}{1mm}
    \caption{\textbf{Computational cost of training the different network components.} We compare the cost of the different network components, namely, the I2I, depth, feature GAN, Laplacian and text module in our method. We report the percentage of the total computational time taken by each of these components. We report that the I2I module dominates the training time. Note that the above methods were trained using 4 GPUs following consistent resolution for all the input images and constant batch size.} 
   \label{tb:computational-cost}%
\vspace{-15pt}
\end{table}%


\mbox{}
\clearpage
\newpage
{\small
\bibliographystyle{ieee_fullname}

}